\documentclass[conference]{IEEEtran}

\usepackage{amsmath,amssymb}
\usepackage{graphicx}
\usepackage{booktabs}
\usepackage{cite}
\usepackage{url}
\usepackage{placeins}
\newcommand{\vect}[1]{\boldsymbol{#1}}

\title{Edge-Native Temporal--Graph Dual-Expert Inference for Dexterous Grasp Classification}

\author{
\IEEEauthorblockN{1\textsuperscript{st} Jiajian Zheng}
\IEEEauthorblockA{
\textit{School of Microelectronics}\\
\textit{Shanghai University}\\
Shanghai, 201800, China\\
24723746@shu.edu.cn
}
\and
\IEEEauthorblockN{2\textsuperscript{nd} Bin Zhang}
\IEEEauthorblockA{
\textit{School of Microelectronics}\\
\textit{Shanghai University}\\
Shanghai, 201800, China\\
zhangbinshu@shu.edu.cn
}
\and
\IEEEauthorblockN{3\textsuperscript{rd} Wen Fan}
\IEEEauthorblockA{
\textit{Department of Bioengineering}\\
\textit{Imperial College London}\\
London SW7 2AZ, United Kingdom\\
w.fan24@imperial.ac.uk
}
\and
\IEEEauthorblockN{4\textsuperscript{th} Yanan Liu\textsuperscript{*}}
\IEEEauthorblockA{
\textit{School of Microelectronics}\\
\textit{Shanghai University}\\
Shanghai, 201800, China\\
yanan\_liu@shu.edu.cn
}
}

\begin{document}
\maketitle

\begin{abstract}
Multi-finger hands contact objects at different times, so a global close command can overload early-contact fingers while others are still moving. Dense tactile arrays address this problem but add hardware, calibration, bandwidth, and computation. We present an edge-native pipeline that uses only built-in joint-angle and pre-use-calibrated actuator-load feedback to recognize the grasped object and predict an independent Stop event for each finger. Four samples and their first differences form a 4-by-20 contact-evolution window. At its core is a heterogeneous dual-expert architecture, termed DualExpert, that assigns complementary tasks to a Temporal Convolutional Network (TCN) and a Finger Graph Filter Network (FGFN). The TCN models short-term contact evolution and supplies all five Stop probabilities, while the FGFN encodes finger topology and contributes only to object classification. Fixed 0.7/0.3 class-probability fusion reaches 93.78\% trajectory-level classification accuracy on 225 held-out trials, improving the standalone TCN and FGFN by 4.00 and 2.22 percentage points, respectively. The TCN Stop path attains 95.25\% F1. The dataset contains 1,500 automatically collected trajectories from five object instances. In Flash deployment, mixed accelerator execution retains 93.78\% classification accuracy and 94.84\% Stop F1 with 2.890~ms mean inference time and 9.172~ms mean sample-to-stop latency. Across 50 physical trials, the system achieves 96.00\% recognition and 98.00\% grasp success while the complete embedded board draws approximately 0.80~W.
\end{abstract}


\section{Introduction}
Dexterous hands conform to varied objects more readily than parallel-jaw grippers, but their fingers often contact the object asynchronously \cite{Kappassov2015Review,Ford2025Shear}. A global closing command may therefore overload one finger while others remain in free motion. Reliable grasping requires both whole-hand object interpretation and independent per-finger stopping.

Dense tactile systems support recognition, slip detection, and closed-loop grasping \cite{Kappassov2015Review,Calandra2018MoreThan,Boonstra2024Slip,Ford2023Gentle}. They can also provide hand-wide representations and learned dexterous policies \cite{Sharma2025PercepSkin,Guzey2024SeeToTouch}, but require additional sensing hardware. Kinesthetic sequences from actuator positions and torques can recognize objects without tactile arrays \cite{Arolovitch2024Kinesthetic}. These results show that actuator histories contain useful object information, but their joint use for object classification, independent per-finger stopping, and fully local embedded execution remains insufficiently explored. This motivates us to use short sequences of native angle and calibrated load feedback for both recognition and stopping.

Low-latency execution is essential when perception directly drives control. Sensor-level systems place sensing, memory, and processing close to the data source to reduce data movement under embedded constraints \cite{Dudek2022SensorLevel,fan2026vision}. Binarized inference and Flash-to-register model swapping further support neural networks under limited memory \cite{Liu2022OnSensor}, while local processing has enabled reactive robot control without transferring full sensor measurements to an external computer \cite{Liu2021AgileNavigation}. Following this principle, we keep the model weights in Flash, stream hand feedback into SRAM, and perform feature construction, inference, thresholding, and persistent finger-state latching entirely on the embedded board.

A further challenge arises because object classification and finger stopping depend on different evidence. Classification benefits from comparing contact patterns across the hand, including the relative responses of neighboring fingers. In contrast, a Stop decision depends more directly on the recent angle and load evolution of each finger. A single shared prediction head may not represent both requirements equally well. We therefore combine a temporal expert with a fixed-topology graph expert for object classification, while retaining the stronger temporal branch for all five Stop predictions. This task-asymmetric design uses graph-based inter-finger evidence where it improves classification without exposing the control output to the weaker graph-based Stop head.

The contributions of this work can be summarized as follows: (1) a task-asymmetric temporal and graph dual-expert model; (2) Flash-resident inference with board-side per-finger control; (3) a leakage-safe dataset of 1,500 trajectories from five object instances; and (4) evaluation across model families, object instances, finger channels, quantization stages, and embedded targets.

\section{Related Work}
\subsection{Temporal tactile and kinesthetic grasp perception}
Temporal contact is more discriminative than an isolated reading for tactile object recognition \cite{Bottcher2021TimeSeries,Kirby2022Dynamic}. Multi-scale and local-global models further aggregate temporal structure \cite{Qian2023TemporalScale,Qian2024LocalGlobal}, while actuator position and torque sequences enable recognition without tactile sensors \cite{Arolovitch2024Kinesthetic}. PercepSkin combines kinematic and distributed tactile histories for reusable hand representations \cite{Sharma2025PercepSkin}. Our method instead uses four frames of native angle and calibrated load feedback and directly couples recognition to per-finger stopping.

Tactile feedback has also supported grasp-stability prediction, slip detection, and adaptive manipulation \cite{Bekiroglu2011Stability,Calandra2018MoreThan,ZapataImpata2019Stability,Boonstra2024Slip,Tao2023AdaptiveFinger}. Recent systems use biomimetic fingertips, visual incentives, or force observations for learned grasp control \cite{Ford2023Gentle,Ford2025Shear,Guzey2024SeeToTouch,Zhang2025TactileRole}. In contrast, our model does not reconstruct tactile images or learn a full manipulation policy. It recognizes one of five object instances and emits five Stop scores during a single closing motion.

\subsection{Topology-aware learning and embedded inference}
Graph models encode persistent spatial relationships in distributed tactile and proprioceptive measurements \cite{Yang2023TacGNN,Defferrard2016ChebNet,Shahidzadeh2024AcTExplore}. General graph operations may map poorly to microcontroller accelerators, so we implement fixed little-to-thumb low- and high-pass filters with standard $1\times1$ convolutions. Depthwise-separable temporal blocks follow efficient mobile and sequence models \cite{Howard2017MobileNets,Bai2018TCN}.

Embedded inference must jointly consider model structure, memory, and runtime \cite{Lin2020MCUNet,Lai2018CMSISNN,Banbury2021MLPerfTiny}. Our weights remain in Flash, measurements and activations occupy SRAM, and decisions remain on the board. Measurements are reported separately for Cortex-M55-only execution and Neural-ART mixed execution, in which unsupported operations are assigned to the on-board Cortex-M55 CPU.

\section{System and Problem Formulation}
\begin{figure*}[!t]
\centering
\includegraphics[width=.98\textwidth,trim=0bp 180bp 0bp 175bp,clip]{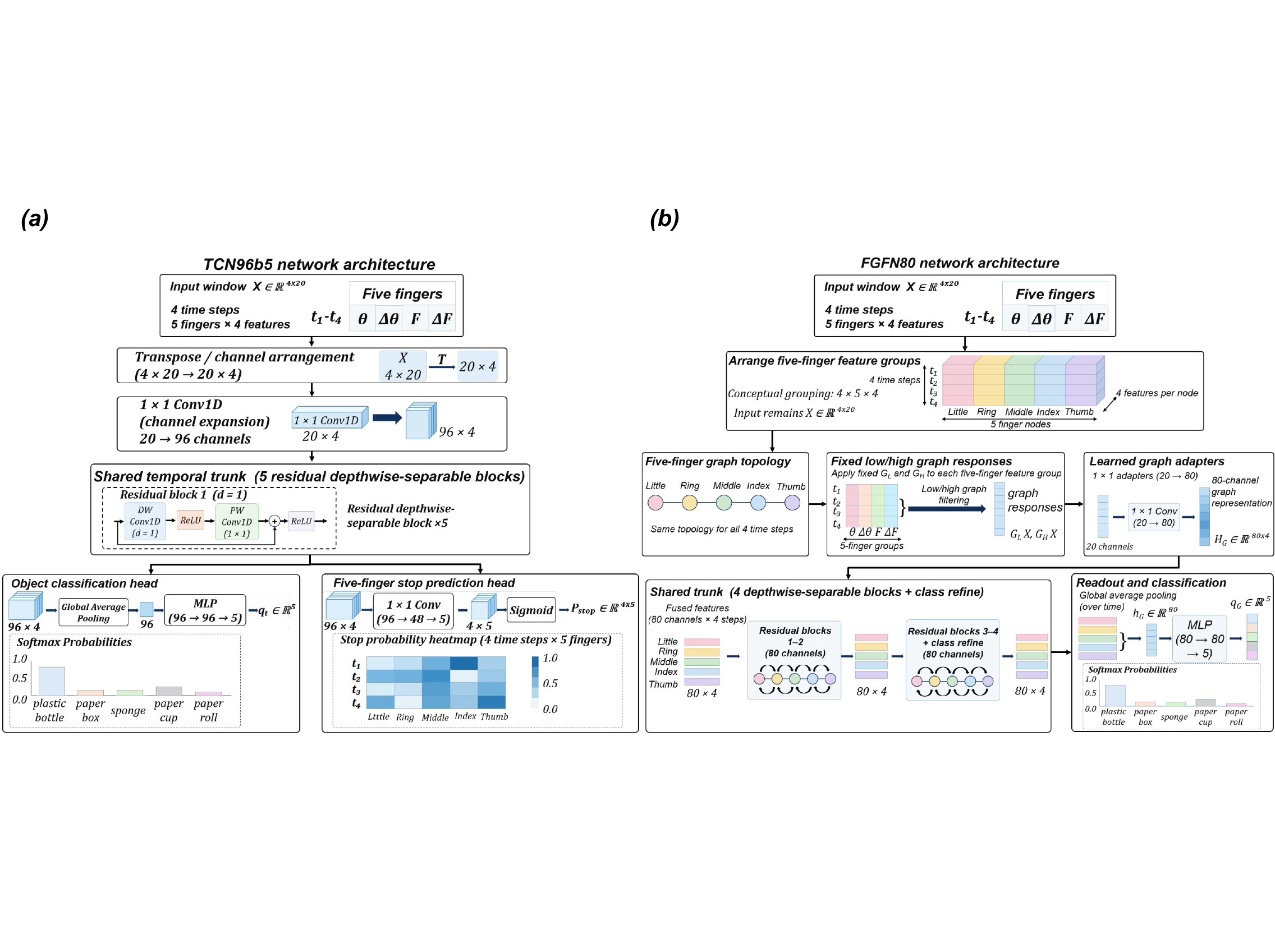}
\caption{Architectures of the complementary experts: (a) TCN96b5 models four-frame contact evolution and predicts object class and five-finger Stop probabilities; (b) FGFN80 injects the fixed finger topology through low/high graph responses for more balanced object classification.}
\label{fig:experts}
\end{figure*}

The system combines sparse hand feedback with local recognition and per-finger stopping. Fig.~\ref{fig:experts} shows the complementary experts. TCN96b5 denotes a 96-channel Temporal Convolutional Network with five residual blocks, while FGFN80 denotes an 80-channel Finger Graph Filter Network. The TCN models contact evolution and stopping; the FGFN exposes neighboring-finger agreement and imbalance for classification.

\subsection{Hardware and feedback path}
The platform combines the RH56 hand, an embedded inference board, and a half-duplex RS-485 link \cite{InspireRH56Manual}. Firmware validates each received frame and uses a second interface only for supervisory commands and logs. The model input contains flexion angles and actuator-load registers ordered from little finger to thumb. Load readings are calibrated before operation but are not SI-unit contact forces, so they are treated as force proxies throughout the pipeline.

\subsection{Four-frame contact-evolution input}
At control instant $t$, let $\vect{a}_t,\vect{f}_t\in\mathbb{R}^{5}$ be the flexion-angle and force-proxy vectors. The board computes
\begin{equation}
\Delta\vect{a}_t=\vect{a}_t-\vect{a}_{t-1},\quad
\Delta\vect{f}_t=\vect{f}_t-\vect{f}_{t-1}.
\end{equation}
Each frame is $\vect{x}_t=[\vect{a}_t,\Delta\vect{a}_t,\vect{f}_t,\Delta\vect{f}_t]\in\mathbb{R}^{20}$. A FIFO forms $\vect{X}_t=[\vect{x}_{t-3},\ldots,\vect{x}_t]\in\mathbb{R}^{4\times20}$, with initial differences set to zero. ONNX \texttt{Sub}/\texttt{Div} nodes embed training-set standardization. The network outputs class probabilities $\vect{p}^{cls}$ and four-frame Stop probabilities $\vect{P}^{stop}$. The controller applies the validation-selected threshold $\tau=0.76$ to the newest row, while trial classification averages window probabilities.

\subsection{Label semantics and supervision}
During incremental closure, the threshold teacher changes $y^{stop}_{t,i}$ from zero to one when finger $i$ reaches its registered load threshold or closed limit. The label then remains latched, while the object label is constant throughout a trial. These labels represent a stopping policy rather than grasp-stability truth; physical success is evaluated separately by lift-and-hold trials. Target angles, thresholds, status words, timestamps, and collection phases are excluded from model inputs.

\section{Heterogeneous Dual-Expert Network}
\begin{figure}[!htbp]
\centering
\includegraphics[width=.98\columnwidth]{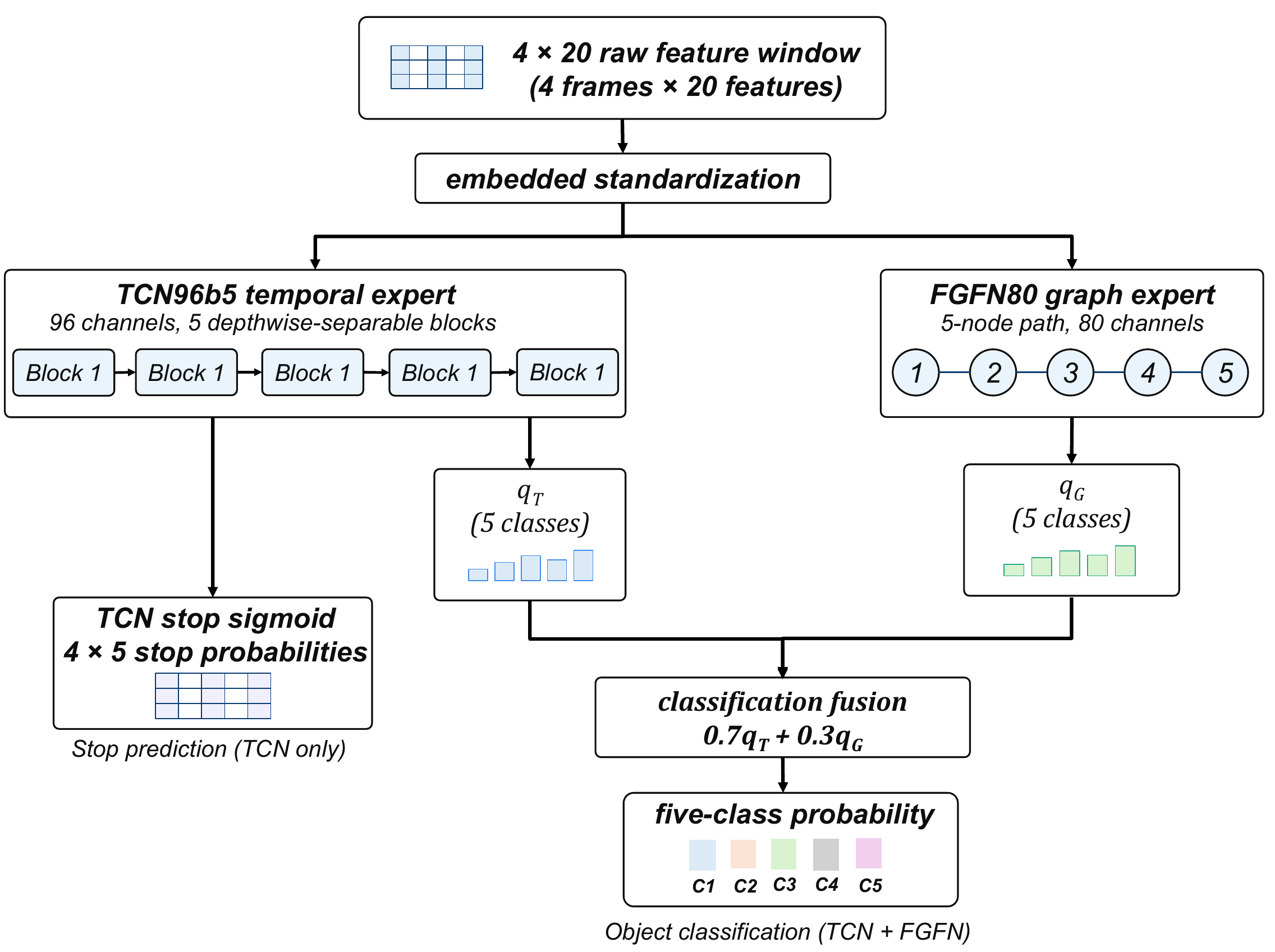}
\caption{Heterogeneous DualExpert. The topology-aware FGFN complements temporal classification, while control-sensitive Stop outputs remain on TCN96b5.}
\label{fig:model}
\end{figure}

Fig.~\ref{fig:model} presents the Heterogeneous Temporal--Graph Dual-Expert, abbreviated as DualExpert. Both experts share the contact window. Their class distributions are fused, while all learned Stop scores come from TCN96b5 because it is substantially stronger on that task. Firmware angle and target limits remain independent completion conditions.

\subsection{Temporal Convolutional Network expert}
As shown in Fig.~\ref{fig:experts}(a), TCN96b5 projects each frame to 96 channels and applies five residual depthwise-separable blocks with $k=3$. Global pooling and a multilayer perceptron produce class logits $\vect{z}_T$; framewise pointwise layers produce Stop logits $\vect{Z}^{stop}_T$. The model contains 65,194 parameters. Keeping its Stop path separate prevents graph-branch classification changes from affecting actuation.

\subsection{Finger Graph Filter Network expert}
As shown in Fig.~\ref{fig:experts}(b), FGFN80 organizes the five fingers as a little-to-thumb path. With adjacency $\vect{A}$ and degree $\vect{D}$, its fixed filters are
\begin{equation}
\vect{G}_{L}=\tfrac{1}{2}\vect{I}+\tfrac{1}{2}\vect{D}^{-1}\vect{A},\qquad
\vect{G}_{H}=2(\vect{I}-\vect{G}_{L}).
\end{equation}
For an interior finger, $G_L$ uses $[0.25,0.5,0.25]$ and $G_H$ uses $[-0.5,1,-0.5]$. Applied to all four feature types, they expose neighboring-finger agreement and local asymmetry. Learned adapters, four depthwise-separable blocks, and a refinement block produce class logits $\vect{z}_G$. FGFN80 contains 57,375 parameters. Its separately trained Stop head is evaluated only as a baseline and is discarded by DualExpert.

\subsection{Class-only fusion and export}
Let $\vect{q}_T=\operatorname{softmax}(\vect{z}_T)$ and $\vect{q}_G=\operatorname{softmax}(\vect{z}_G)$. DualExpert computes
\begin{equation}
p^{cls}_c=\alpha q_{T,c}+(1-\alpha)q_{G,c},
\end{equation}
with $\alpha=0.7$ and no class bias. The learned Stop output follows a separate temporal path:
\begin{equation}
\vect{P}^{stop}=\operatorname{sigmoid}(\vect{Z}^{stop}_T).
\end{equation}
This asymmetric fusion adds topology-aware class evidence without degrading the stronger TCN Stop output. Both experts use the same trajectory split and training-set standardization. The fusion weight is selected on the validation set and fixed before test evaluation.

Export embeds standardization, Softmax, fusion, and the TCN Sigmoid. DualExpert contains 122,569 parameters. QDQ conversion uses representative training windows; Cortex-M55 uses asymmetric QUInt8, while Neural-ART uses symmetric QInt8 with percentile calibration. Each target is evaluated against its matching computer-side INT8 reference.

\section{Embedded Deployment and Control}
\subsection{Target-specific deployment}
ST Edge AI stores model weights and operators with the firmware in Flash. At runtime, each validated RS-485 sample enters SRAM, updates the four-frame buffer, and forms the normalized $4\times20$ input. The Cortex-M55-only or Neural-ART mixed-execution path writes class and Stop probabilities back to SRAM for immediate finger-latch updates, without returning raw windows to the computer.

The Cortex-M55-only build uses 178.08~KiB of weights, 8.46~KiB of peak activations, and 405,516 MACs per window. The Neural-ART mixed-execution build stores a 157.61-KiB model blob, uses 5.77~KiB of peak activations, and reports 470,599 compiler-counted MACs. Unsupported normalization, fusion, Softmax, and Sigmoid operations cause 32 of 88 compiled execution epochs to run on the on-board Cortex-M55 CPU, while the remaining epochs are executed by the Neural-ART accelerator. We therefore refer to this configuration as mixed execution.

\subsection{Latched incremental closure}
After four-sample warm-up, a Stop event latches when $p^{stop}_{t,i}\geq0.76$. Finger $i$ then holds its measured angle while the others continue closing. Mechanical limits and an iteration guard remain independent safety conditions. The command is
\begin{equation}
u_{t+1,i}=\begin{cases}
a_{t,i}, & s_{t,i}=1,\\
\max(u_{t,i}-\delta_u,0), & s_{t,i}=0,
\end{cases}
\end{equation}
where smaller RH56 units mean closure, $\delta_u$ is the closing increment, and $s_{t,i}$ is persistent. Errors abort motion, and online scheduling retains only the latest four frames. Mean end-to-end sample-to-stop latency is 9.172~ms, including acquisition, preprocessing, inference, latching, and command transmission.

\begin{figure}[!htbp]
\centering
\includegraphics[width=.98\columnwidth]{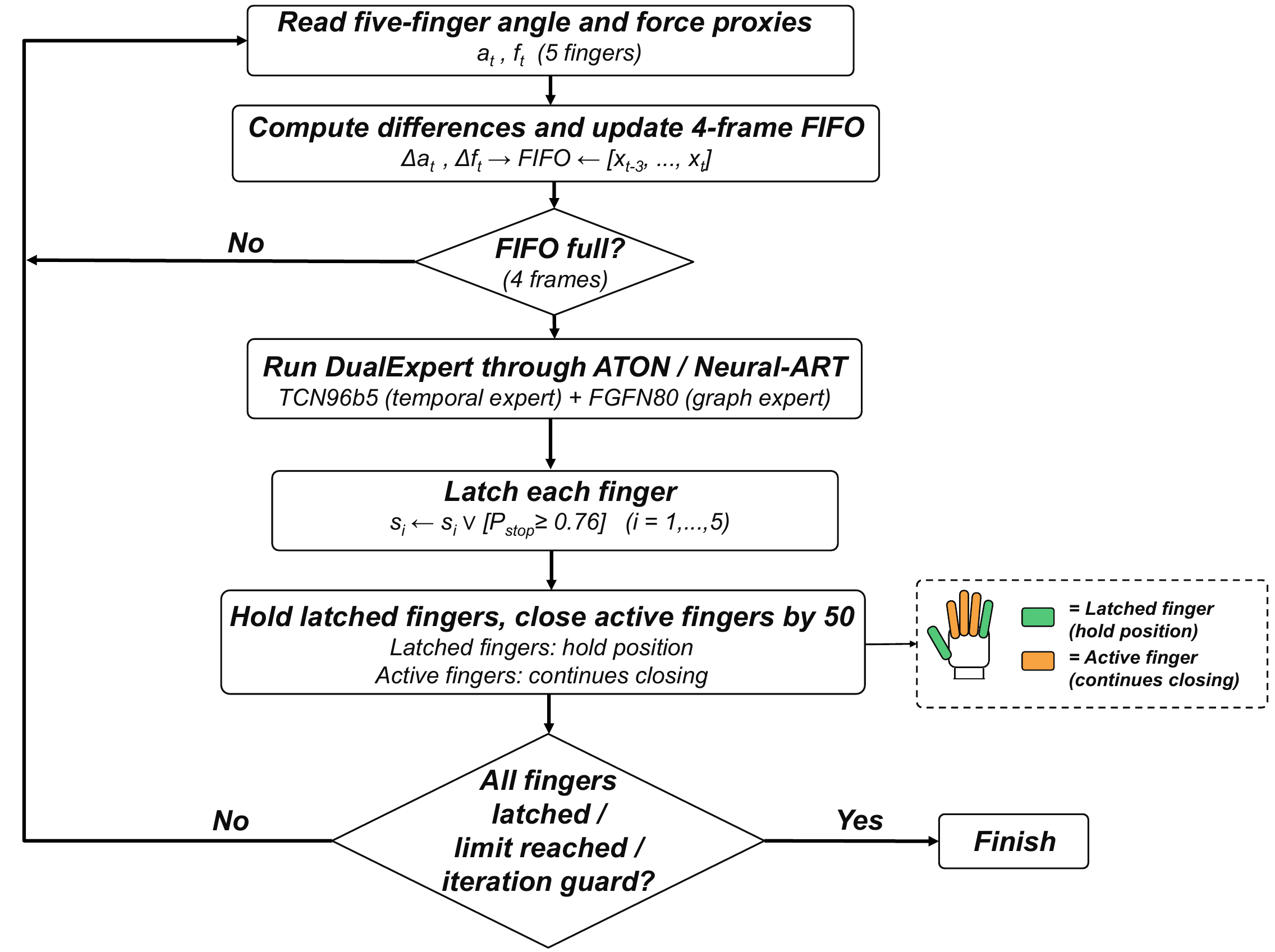}
\caption{On-board per-finger grasp flow. Measured feedback forms each new contact frame; model-triggered Stop states are monotonic while other fingers continue incremental closure.}
\label{fig:control}
\end{figure}

Fig.~\ref{fig:control} summarizes how sensing, inference, and per-finger actuation form one control cycle.

\section{Experiments}
\subsection{Protocol and implementation}
\begin{figure}[!htbp]
\centering
\includegraphics[width=.98\columnwidth]{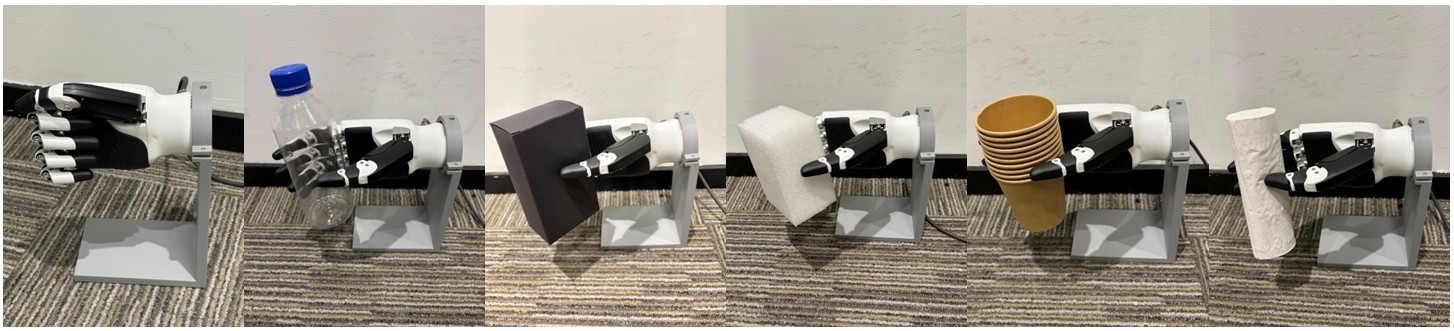}
\caption{Experimental platform and evaluated instances. From left to right: RH56 hand, plastic bottle, paper box, sponge, paper cup, and paper roll.}
\label{fig:setup}
\end{figure}

\begin{figure}[!htbp]
\centering
\includegraphics[width=.98\columnwidth,trim=5bp 220bp 10bp 160bp,clip]{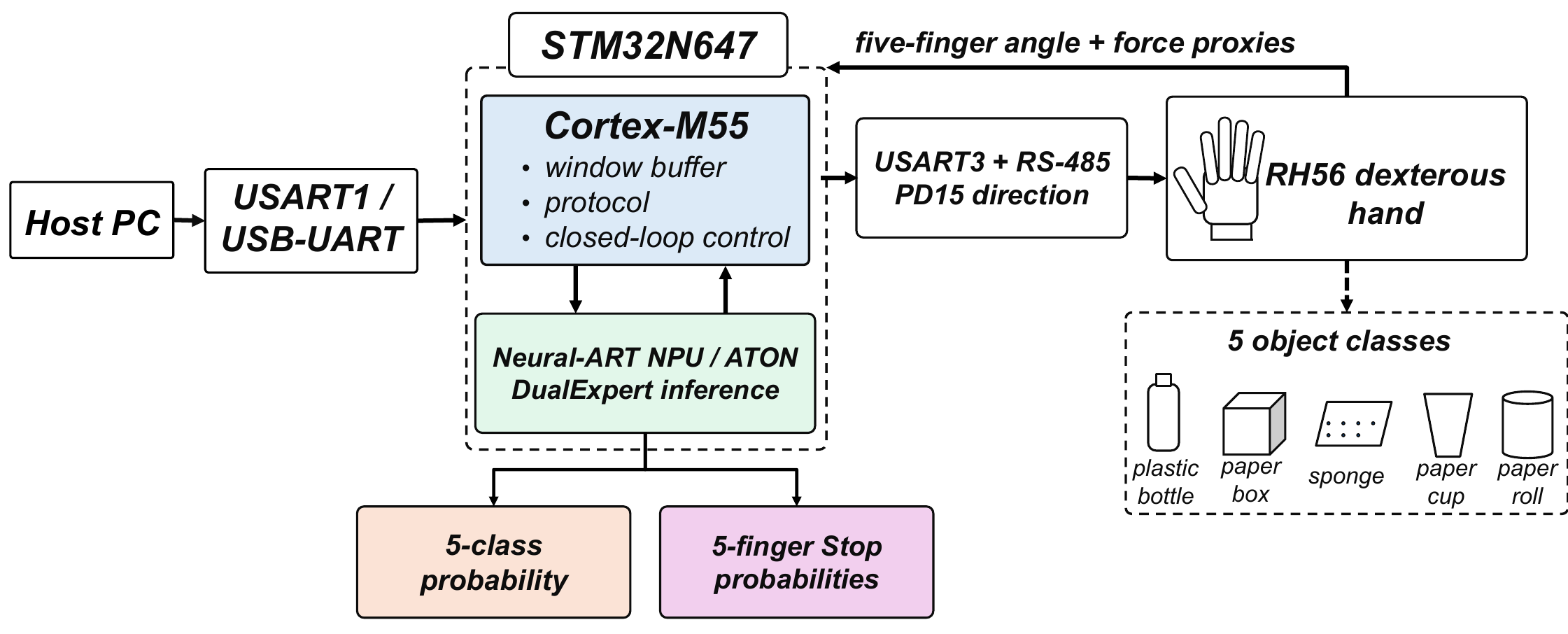}
\caption{Board-side streaming architecture used in the deployment experiments. Samples enter temporary SRAM once; feature construction and inference execute locally with immutable model parameters stored in Flash.}
\label{fig:system}
\end{figure}

Experiments evaluate recognition, per-finger stopping, expert complementarity, and embedded execution using the platform and five object instances shown in Fig.~\ref{fig:setup}. Deployment uses an STM32N647 on the ATK-DNN647 board and evaluates both Cortex-M55-only execution and Neural-ART mixed execution \cite{AlientekDNN647,ST2025N6AI}. The measured board-side data path is shown in Fig.~\ref{fig:system}.

The acquisition application performs repeatable close-record-open cycles over RS-485 and saves a trial only after complete five-finger feedback and a final synchronized read. The dataset contains 300 trajectories for each object in Fig.~\ref{fig:setup}, collected with the object randomly placed at different in-plane orientations, resulting in 1,500 trajectories in total. Each object class is stratified into 210 training, 45 validation, and 45 test trajectories before window extraction, preventing windows from the same trial from appearing in different splits. These partitions contain 303,956, 66,019, and 64,628 windows, respectively. Stop-positive test samples from little finger to thumb number 225, 9,288, 16,145, 22,953, and 19,911. Standardization is fitted only on training data.

All component networks use weighted cross-entropy and per-finger weighted binary cross-entropy,
$\mathcal{L}=1.2\mathcal{L}_{cls}+\mathcal{L}_{stop}$.
Training uses AdamW with learning rate $10^{-3}$, weight decay $10^{-4}$, cosine annealing, batch size 16,384, and at most 50 epochs. DualExpert combines independently trained checkpoints with fixed 0.7/0.3 class fusion; its 0.76 Stop threshold is selected on validation data.

Classification is measured per complete test trajectory by averaging window probabilities. Stop precision, recall, and F1 are micro-aggregated across all test-window finger decisions. Latency is measured per $4\times20$ window.

\subsection{Seven-model comparison}
Table~\ref{tab:modelbench} reports performance on the held-out test set. The conventional MLP, 1D CNN, GRU, and LSTM baselines obtain lower classification and Stop scores and show an uneven balance between the two tasks. TCN96b5 substantially improves Stop prediction but remains less consistent across object classes, whereas FGFN80 provides more balanced classification at the cost of weaker Stop performance. DualExpert combines these complementary strengths, achieving 93.78\% classification accuracy and 93.86\% macro F1, which improves on TCN96b5 and FGFN80 by 4.00 and 2.22 accuracy points, respectively. Its Stop metrics equal those of TCN96b5 because the temporal path is used unchanged. Computer-side latency is measured on an RTX 5060 Laptop GPU.

DualExpert processes 432.1 windows/s on the computer, exceeding the 100-Hz sampling rate used for data acquisition. Once inference throughput exceeds this input rate, a faster model does not increase the effective decision update rate. DualExpert therefore uses the available compute margin to improve joint performance without reducing the system update rate.

\begin{table}[t]
\caption{Held-out test performance and computational cost}
\label{tab:modelbench}
\centering
\scriptsize
\setlength{\tabcolsep}{2.2pt}
\resizebox{\columnwidth}{!}{%
\begin{tabular}{lrrrrrr}
\toprule
Model & Class acc. (\%) & Macro F1 (\%) & Stop F1 (\%) & Params & Mean (ms) & P95 (ms)\\
\midrule
MLP & 68.00 & 67.25 & 78.02 & 30,105 & 0.261 & 0.481\\
1D CNN & 77.78 & 77.31 & 82.64 & 29,258 & 0.648 & 1.126\\
GRU & 80.00 & 79.59 & 82.64 & 46,282 & 0.301 & 0.445\\
LSTM & 84.44 & 84.32 & 82.46 & 60,106 & 0.405 & 0.758\\
TCN96b5 & 89.78 & 89.62 & \textbf{95.25} & 65,194 & 1.490 & 2.580\\
FGFN80 & 91.56 & 91.55 & 81.92 & 57,375 & 1.223 & 1.792\\
DualExpert & \textbf{93.78} & \textbf{93.86} & \textbf{95.25} & 122,569 & 2.314 & 3.296\\
\bottomrule
\end{tabular}
}
\end{table}

\begin{figure}[t]
\centering
\includegraphics[width=.96\columnwidth]{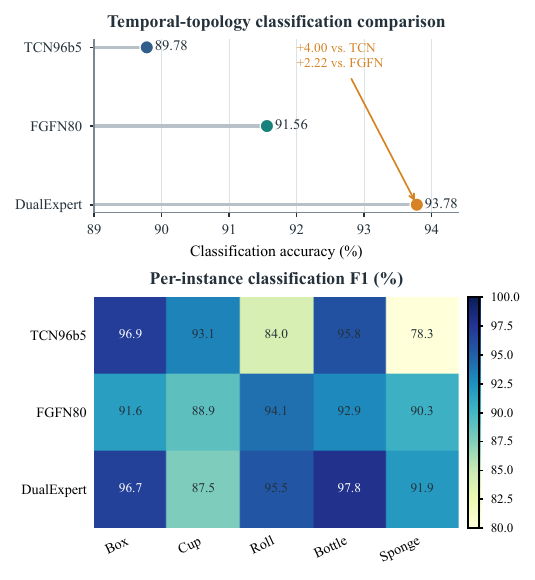}
\caption{Classification complementarity of TCN96b5, FGFN80, and DualExpert. The upper panel reports classification accuracy; the lower heatmap reports per-instance class F1.}
\label{fig:advantage}
\end{figure}

Fig.~\ref{fig:advantage} shows complementary class behavior. TCN96b5 performs strongly on the bottle, box, and cup but falls to 83.96\% and 78.35\% F1 on the roll and sponge. FGFN80 remains within 88.88--94.10\%, consistent with more balanced inter-finger evidence. Fusion obtains the highest aggregate accuracy and improves the bottle, roll, and sponge results, although it is not best on every class.

\subsection{Per-object and per-finger stopping}
Per-object DualExpert Stop F1 ranges from 93.67\% to 96.24\%. Sponge has the highest missed-stop rate at 5.65\%, while paper roll has a 6.74\% false-stop ratio.

\begin{figure}[t]
\centering
\includegraphics[width=.96\columnwidth]{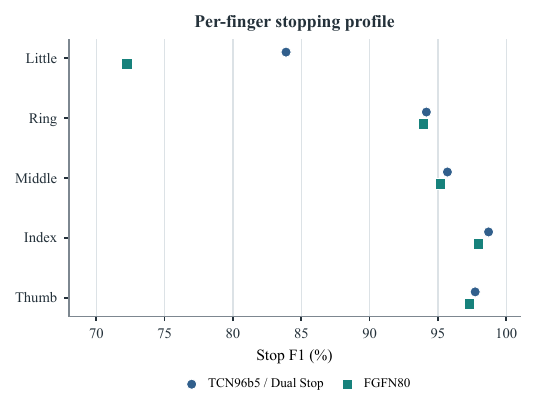}
\caption{Per-finger Stop F1 for the TCN96b5 path used unchanged by DualExpert and the independently evaluated FGFN80 Stop head.}
\label{fig:fingerstop}
\end{figure}

Fig.~\ref{fig:fingerstop} shows that little-finger F1 is 83.88\%, compared with 94.15--98.70\% for the other fingers. Its test set contains only 225 positive samples, so aggregate Stop F1 should be read together with channel-level results. FGFN80's lower Stop F1 supports retaining the TCN path without fusion.

\subsection{Conversion and embedded replay}
Table~\ref{tab:deploy} evaluates the final DualExpert graph. ONNX FP32 agrees with PyTorch on 99.9923\% of window classifications and 99.9984\% of thresholded Stop outputs across 64,628 test windows. Board metrics are measured after compiling the same graph into Flash.

\begin{table}[t]
\caption{Computer and target execution summary}
\label{tab:deploy}
\centering
\scriptsize
\setlength{\tabcolsep}{2.8pt}
\begin{tabular}{lrrr}
\toprule
Execution & Class acc. (\%) & Stop F1 (\%) & Mean (ms)\\
\midrule
PyTorch & 93.78 & 95.25 & 2.314\\
ONNX FP32 & 93.78 & 95.16 & 0.207\\
Cortex-M55 only & 93.33 & 94.17 & 3.949\\
Neural-ART mixed & 93.78 & 94.84 & 2.890\\
\bottomrule
\end{tabular}
\end{table}

Cortex-M55-only execution reaches 253.3 windows/s, while Neural-ART mixed execution reaches 346.1 windows/s. Both exceed the 100-Hz sampling rate. Neural-ART mixed execution reduces mean inference latency by 26.80\%, with 32 of 88 compiled execution epochs running on the on-board Cortex-M55 CPU.

\subsection{Physical closed-loop evaluation}
DualExpert is tested ten times per object in Fig.~\ref{fig:setup}. Success requires the object to remain lifted for 10 seconds without visible slip, dropping, or exceeding the registered load limit. Table~\ref{tab:grasp} reports object dimensions, mass, the number of correct recognitions, and the number of successful grasps. Box and sponge sizes follow $L\times W\times H$; round objects use diameter and height.

\begin{table}[!htbp]
\caption{Physical objects and closed-loop outcomes}
\label{tab:grasp}
\centering
\scriptsize
\setlength{\tabcolsep}{2.2pt}
\resizebox{\columnwidth}{!}{%
\begin{tabular}{lccrrr}
\toprule
Object & Size (mm) & Mass (g) & Trials &
\shortstack{Correct\\recognition} &
\shortstack{Successful\\grasps}\\
\midrule
Plastic bottle & $\varnothing59\!\times\!161$ & 20.0 & 10 & 10 & 10\\
Paper box & $76.8\!\times\!52.5\!\times\!136.4$ & 18.5 & 10 & 10 & 10\\
Sponge & $72.0\!\times\!56.1\!\times\!104.0$ & 4.6 & 10 & 9 & 10\\
Paper cup & $\varnothing72.4/\varnothing53.5\!\times\!123.3$ & 41.9 & 10 & 10 & 10\\
Paper roll & $\varnothing40.6\!\times\!150.2$ & 33.4 & 10 & 9 & 9\\
Overall & -- & -- & 50 & 48 & 49\\
\bottomrule
\end{tabular}
}
\end{table}

Mean sample-to-stop latency is 9.172~ms, including communication, preprocessing, inference, and latching. The complete embedded board consumed 0.74~W at idle and 0.80~W during sustained Neural-ART mixed execution. Under FP32 inference, the RTX 5060 Laptop GPU alone drew 30.80~W against a 17.07-W idle baseline. These measurements show that board-side inference requires substantially less operating power than computer-side GPU inference. Host CPU, memory, display, and power-supply losses are excluded from the computer-side measurement.

\section{Discussion}
Table~\ref{tab:modelbench} and Fig.~\ref{fig:advantage} support the task-asymmetric design: TCN96b5 provides stronger Stop prediction, whereas FGFN80 gives more consistent classification across object instances. Their class-probability fusion improves overall classification while retaining the TCN Stop path. Neural-ART mixed execution reaches 346.1 windows/s, exceeding the 100-Hz input rate and meeting the real-time requirement without computer-side inference.

With one physical instance per label, the evaluation measures instance recognition rather than category-level generalization. Class-specific Stop thresholds may also influence the trajectories. Future work should include additional instances, out-of-plane poses, wider geometric and material variation, and repeated power and latency measurements under controlled conditions.

\section{Conclusion}
DualExpert combines temporal contact evolution, fixed finger topology, and board-side control, reaching 93.78\% classification accuracy and 95.25\% Stop F1 on 225 held-out trajectories. Neural-ART mixed execution retains the same classification accuracy with 94.84\% Stop F1 at 2.890~ms mean inference time. Physical trials achieve 98.00\% grasp success, 9.172~ms sample-to-stop latency, and approximately 0.80~W complete-board power.

\section*{Acknowledgment}
The authors thank Beijing Inspire-Robots Technology Co., Ltd. (INSPIRE-ROBOTS) for hardware and technical support.

\FloatBarrier
\bibliographystyle{IEEEtran}
\bibliography{references}

\end{document}